%% file: aaai23.tex
\documentclass[letterpaper]{article} 
\usepackage{aaai23}  
\usepackage{times}  
\usepackage{helvet}  
\usepackage{courier}  
\usepackage[hyphens]{url}  
\usepackage{graphicx} 
\usepackage{natbib}  
\usepackage{caption} 
\usepackage[table,dvipsnames]{xcolor}
\usepackage{graphicx}

\makeatletter
\newcommand{\thickhline}{%
    \noalign {\ifnum 0=`}\fi \hrule height 1pt
    \futurelet \reserved@a \@xhline
}
\newcolumntype{"}{@{\hskip\tabcolsep\vrule width 1pt\hskip\tabcolsep}}
\makeatother

\usepackage{algorithm}
\usepackage{algpseudocode}
\algrenewcommand\algorithmicrequire{\textbf{Input:}}
\algrenewcommand\algorithmicensure{\textbf{Output:}}
\algnewcommand{\NULL}{\textsc{null}}

\usepackage{multirow}
\usepackage{amsmath}
\newcommand{\norm}[1]{\left\lVert#1\right\rVert}
\usepackage{booktabs}
\usepackage{amsthm}

\newtheorem{property}{Property}
\usepackage{amssymb}
\usepackage{makecell}
\usepackage[toc,page]{appendix}
\usepackage{xr}
\usepackage{subcaption}
\usepackage{arydshln}

\DeclareMathOperator*{\argmax}{arg\,max}
\DeclareMathOperator*{\argmin}{arg\,min}

\usepackage{newfloat}
\usepackage{listings}
\DeclareCaptionStyle{ruled}{labelfont=normalfont,labelsep=colon,strut=off} 
\floatstyle{ruled}
\newfloat{listing}{tb}{lst}{}
\floatname{listing}{Listing}
\title{Graph Component Contrastive Learning for Concept Relatedness Estimation}
\author{
    Yueen Ma\textsuperscript{\rm 1}, Zixing Song\textsuperscript{\rm 1}, Xuming Hu\textsuperscript{\rm 2}, Jingjing Li\textsuperscript{\rm 1}, Yifei Zhang\textsuperscript{\rm 1}, Irwin King\textsuperscript{\rm 1}\\
}
\affiliations{
    \textsuperscript{\rm 1}The Chinese University of Hong Kong,
    \textsuperscript{\rm 2}Tsinghua University\\

    \texttt{\{yema21, zxsong, lijj, yfzhang, king\}@cse.cuhk.edu.hk\\
    hxm19@mails.tsinghua.edu.cn}
}

\usepackage{bibentry}

\begin{document}

\maketitle

\begin{abstract}
Concept relatedness estimation (CRE) aims to determine whether two given concepts are related. Existing methods only consider the pairwise relationship between concepts, while overlooking the higher-order relationship that could be encoded in a concept-level graph structure. We discover that this underlying graph satisfies a set of intrinsic properties of CRE, including \textit{reflexivity}, \textit{commutativity}, and \textit{transitivity}. In this paper, we formalize the CRE properties and introduce a graph structure named ConcreteGraph. To address the data scarcity issue in CRE, we introduce a novel data augmentation approach to sample new concept pairs from the graph. As it is intractable for data augmentation to fully capture the structural information of the ConcreteGraph due to a large amount of potential concept pairs, we further introduce a novel Graph Component Contrastive Learning framework to implicitly learn the complete structure of the ConcreteGraph. Empirical results on three datasets show significant improvement over the state-of-the-art model. Detailed ablation studies demonstrate that our proposed approach can effectively capture the high-order relationship among concepts.
\end{abstract}

\input{sections/intro}

\input{sections/related}

\input{sections/method}

\input{sections/experiments}

\section{Conclusion}
Concept relatedness estimation is an emerging task that has a wide range of applications. To the best of our knowledge, we are the first to discover the concept-level graph structure that unveils the high-order relationship among concepts. We name it ConcreteGraph and develop a novel data augmentation method based on it. However, because data augmentation cannot capture the global structural information of the ConcreteGraph, we integrate a novel Graph Component Contrastive Learning (GCCL) framework to encode the complete graph structure. Experimental results show that the data augmentation method can improve the performance of the Transformer models by approximately $1\%$, whereas the improvement from the GCCL framework can be up to $\sim 3\%$. In some cases, data augmentation can complement GCCL and further enhance performance. Detailed analysis shows that GCCL better organizes the embedding space where related concepts are close while unrelated concepts are separated. We also conduct an experiment to show that it is difficult for the ConcreteGraph-based data augmentation method to provide as much performance benefit as GCCL by simply sampling more concept pairs, due to the degradation of the data quality caused by the long path length.



\section{Acknowledgements}
The work described here was partially supported by grants from the National Key Research and Development Program of China (No. 2018AAA0100204) and from the Research Grants Council of the Hong Kong Special Administrative Region, China (CUHK 2410021, Research Impact Fund, No. R5034-18).


\input{aaai23.bbl}

\end{document}

%% file: sections/intro.tex
\section{Introduction}

Concept relatedness estimation (CRE) is the task of determining whether two concepts are related. In CRE, a concept can be a Wikipedia entry, a news article, a social media post, etc. Table~\ref{tbl:cre_example} includes a pair of related concepts and an unrelated concept. In this example, when given the first two concepts, ``Open-source software'' and ``GNU General Public License'', the goal of CRE is to label them as a related pair; but for ``Open-source software'' and ``Landscape architecture'', one should label them as unrelated. 

Existing CRE methods \cite{DBLP:conf/acl/LiuNWGHLX19} only consider the provided concept pairs, which are low-order pairwise relationships. However, we discover that there exist higher-order relationships in CRE that can be encoded by a concept-level graph structure, where the original pairs correspond to immediate neighbors \cite{DBLP:conf/lrec/Ein-DorHKLMRSS18, DBLP:conf/acl/LiuNWGHLX19}. The higher-order relationships are manifested as the multi-hop neighbors in the graph, which are validated by three types of intrinsic properties of CRE: \textit{reflexivity}, \textit{commutativity}, and \textit{transitivity}. To construct the graph, we treat each concept as a node and add edges between any related concept pairs. In this paper, we name it ConcreteGraph (\textbf{\underline{Conc}}ept \textbf{\underline{re}}la\textbf{\underline{te}}dness \textbf{\underline{Graph}}). It enables us to obtain new concept pairs that are not limited to low-order relationships. For example, one of the \textit{transitivity} properties states that if concepts $(x_A, x_B)$ are related and concepts $(x_B, x_C)$ are related, then we can extract a new related concept pair $(x_A, x_C)$, where $x_C$ is a 2-hop neighbor of $x_A$. Therefore, when we sample new concept pairs from the ConcreteGraph, the CRE properties are implicitly utilized.

\begin{table}[t]
\begin{center}
    \resizebox{1\columnwidth}{!}{
        \begin{tabular}{p{1.6cm}|p{8cm}}
        \toprule
            Relatedness & Concept \\
         \hline
         \multirow{2}{*}[-4em]{Related} & \textbf{Open-source software} (OSS) is computer software that is released under a license in which the copyright holder grants users the rights to use, study, change, and distribute the software and its source code to anyone and for any purpose $\cdots$ \\
         \cline{2-2}
          & The \textbf{GNU General Public License} (GNU GPL or simply GPL) is a series of widely used free software licenses that guarantee end users the freedom to run, study, share, and modify the software $\cdots$\\
         \hline
         \multirow{1}{*}[-1.25em]{Unrelated} & \textbf{Landscape architecture} is the design of outdoor areas, landmarks, and structures to achieve environmental, social-behavioural, or aesthetic outcomes $\cdots$\\
        \bottomrule
        \end{tabular}
    }
\caption{Examples of concept relatedness estimation from the WORD dataset.}
\label{tbl:cre_example}
\end{center}
\vspace{-20pt}
\end{table}

\begin{figure}[t]
    \centering
    \includegraphics[width=1\columnwidth,trim={1.1cm 0.5cm 0.5cm 0.5cm},clip]{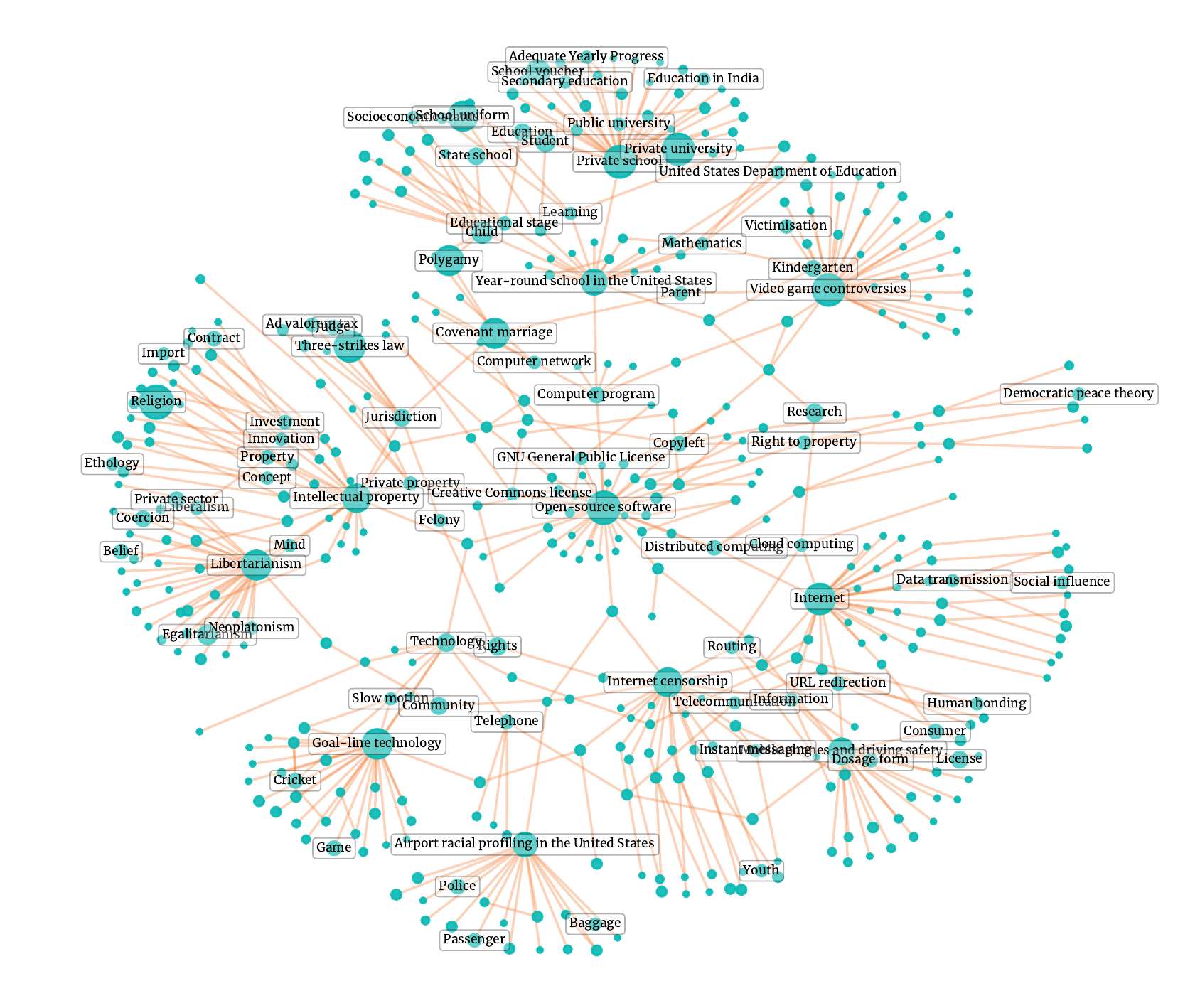}
    \caption{The 4-hop neighborhood of the concept ``Open-source software'' in the ConcreteGraph.}
    \label{fig:concretegraph_example}
\vspace{-15pt}
\end{figure}

The main challenge is how to take full advantage of the structural information of the ConcreteGraph. The most straightforward approach is to explicitly add new concept pairs to the dataset, which can be seen as a data augmentation method for alleviating the data scarcity issue. However, this approach exhibits a problem: there is an excessive amount of potential concept pairs. Assuming that we extract all possible related concept pairs by adding edges between any two connected concepts in the ConcreteGraph, we essentially turn every component of the ConcreteGraph into a complete subgraph. Because the edges in the original ConcreteGraph are sparse, the number of new related concept pairs grows quadratically with the number of concepts. It would become impractical for a model to learn all of them and, thus, we must only sample a subset of all possible concept pairs. If we sample completely at random, the quality of the new concept pairs cannot be guaranteed. That is, when the path length between two concepts is long, the quality of their relationship tends to be low because the probability of the existence of a noisy edge on the path becomes high. This phenomenon is later confirmed by our experiments. A simple solution is to extract concept pairs from the local $k$-hop neighborhood. For instance, we only look for new pairs in the 4-hop neighborhood of a concept, as shown in Figure~\ref{fig:concretegraph_example}. But this limits the ConcreteGraph-based data augmentation method to contain only the local graph structure.


To this end, we introduce a novel Graph Component Contrastive Learning (GCCL) framework that can implicitly capture the complete structural information of the ConcreteGraph. Rather than explicitly learning new concept pairs, the objective of GCCL is that a concept treats its own component as positive, whereas it forms negative pairs with all other components. The representation of a concept or a component is achieved using a shared backbone encoder. Thus, GCCL gives the encoder access to both local and global structural information, which would be intractable if we used the ConcreteGraph-based data augmentation method. GCCL is inspired by Prototypical Contrastive Learning (PCL) \cite{DBLP:conf/iclr/0001ZXH21}, but the difference is that it does not use $k$-means to find clusters. Each component of the ConcreteGraph corresponds to a cluster, and therefore, components in GCCL translate to prototypes in PCL.

Specifically, GCCL involves three main steps: (1) we use a Transformer model as the backbone encoder to embed all concepts and components before each epoch; (2) for each concept, we train the encoder to distinguish the concept's own component from all other components, which is defined to be the GC-NCE objective; (3) along with the main GC-NCE target, we also use a momentum encoder as in the Momentum Contrast (MoCo) framework \cite{DBLP:conf/cvpr/He0WXG20, DBLP:journals/corr/abs-2003-04297} to optimize an InfoNCE loss \cite{DBLP:journals/corr/abs-1807-03748}. This InfoNCE loss contrasts a concept with other concepts. It is added to the main GC-NCE target to preserve the local smoothness of the overall GCCL loss function \cite{DBLP:conf/iclr/0001ZXH21}. We conduct comprehensive experiments with three different Transformer models on three datasets. The empirical results show that the GCCL framework is capable of significantly improving their performance and, when combined with the ConcreteGraph-based data augmentation method, it can sometimes bring even more improvement. We also conduct detailed ablation studies to show the effectiveness of each module in our method. Our code is available on Github\footnote{Github: \url{https://github.com/Panmani/GCCL}}.

The main contributions of our paper are as follows:
\begin{itemize}
  \item For CRE, we formally summarize its intrinsic properties that can be categorized into three types: \textit{reflexivity}, \textit{commutativity}, and \textit{transitivity}. On their basis, we construct a concept-level graph structure, ConcreteGraph, which encodes not only the provided concept pairs but also the higher-order relationships between concepts. To the best of our knowledge, we are the first to find such a graph structure for the CRE task. 

  \item We propose a novel Graph Component Contrastive Learning (GCCL) framework for taking full advantage of the local and global structural information of the ConcreteGraph. GCCL can be complemented by a novel ConcreteGraph-based data augmentation method that explicitly provides local neighborhood information.

  \item  Our method significantly improves over the state-of-the-art Concept Interaction Graph (CIG) model \cite{DBLP:conf/acl/LiuNWGHLX19} on two datasets, CNSE and CNSS. We are also the first to apply deep neural networks to the WORD dataset \cite{DBLP:conf/lrec/Ein-DorHKLMRSS18}.
\end{itemize}


\begin{figure*}
    \centering
    \includegraphics[width=1.8\columnwidth,trim={5.5cm 7cm 5.5cm 8.5cm},clip]{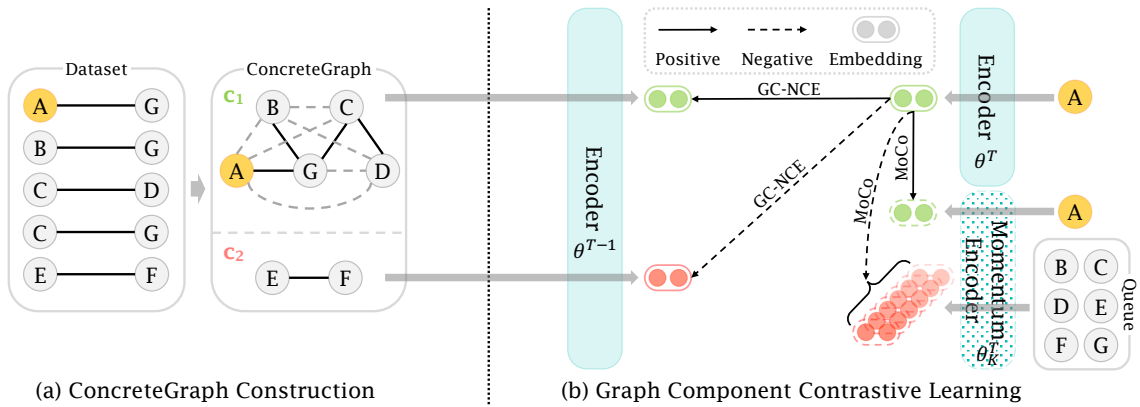}
    \caption{Overview of our proposed framework for concept relatedness estimation. (a) After collecting the concept pairs into the ConcreteGraph, the concepts are divided into multiple components, e.g., $c_1$ and $c_2$. For simplicity, concepts are represented with their indices, e.g., the node with index A represents concept $x_A$; edges connect related concepts: solid black edges are true annotations, and dashed gray edges correspond to new related concept pairs from ConcreteGraph-based data augmentation; disconnected pairs are unrelated ones, such as $(x_A, x_F)$. (b) Under the Graph Component Contrastive Learning (GCCL) framework, the GC-NCE objective aims to distinguish each concept's own component from other components (in this example, the concept is $x_A$, highlighted in yellow); the MoCo objective is to contrast the momentum embedding of each concept with the momentum embeddings of a queue of other concepts. The positive embeddings are colored in green and the negative ones are in red. $\theta^{T-1}$ and $\theta^{T}$ are both the parameters of the main encoder but from two different epochs; $\theta^{T}_{K}$ is the parameters of the momentum encoder from the MoCo framework. }
    \label{fig:overview}
\end{figure*}

%% file: sections/related.tex
\section{Related Work}
\subsection{Concept Relatedness Estimation} 
The concept relatedness estimation (CRE) task stemmed from the concept similarity matching (CSM) task in the area of formal concept analysis (FCA). In FCA, a concept is formally defined as a pair of sets: a set of objects and a set of attributes in a given domain \cite{formica2006ontology}. Methods for assessing concept similarity include ontology-based methods \cite{formica2006ontology, formica2008concept}, Tversky’s-Ratio-based methods \cite{lombardi2006concept}, rough-set-based methods \cite{WANG2008842}, and semantic-distance-based methods \cite{DBLP:conf/skg/GeQ08, li2011method}. The definition for concepts in FCA is not suitable for the CRE task because CRE concepts are long text documents and, thus, the CSM methods cannot be applied to the CRE task. 
Inspired by the giant success of introducing deep neural networks into natural language processing applications~\cite{li2019improving, gao2020discern,sun2022unified, li2020unsupervised}, we adopt Transformer models \cite{DBLP:conf/nips/VaswaniSPUJGKP17} to address the CRE task.

CRE is also related to tasks such as semantic textual similarity \cite{cer2017semeval, zhang2019doc2hash,zhang2020discrete}, text similarity \cite{Thijs2019ParagraphbasedIA}, text relatedness \cite{DBLP:journals/corr/TsatsaronisVV14}, text matching \cite{10.1145/3308558.3313707}, and text classification \cite{hu2020selfore,hu2021semi,hu2021gradient,liu2022hierarchical, li2022text}. The current state-of-the-art model for semantic textual similarity is XLNet \cite{DBLP:conf/nips/YangDYCSL19}, but there is still no work that applies deep neural networks to the WORD dataset. With the advancement of Graph Neural Networks (GNNs) \cite{DBLP:conf/aaai/SongK22, DBLP:conf/kdd/SongZK22,DBLP:conf/cikm/SongMZK21,9737635, zhang2022graph}, \citet{DBLP:conf/acl/LiuNWGHLX19} introduced the Concept Interaction Graph (CIG) method to match news article pairs along with two new datasets, CNSE and CNSS. CIG is the current state-of-the-art model on these two datasets.

CRE can play an important role in a wide range of applications, such as information retrieval \cite{6228205, Teevan2011TwitterSearchAC}, document clustering \cite{ASWANIKUMAR20102696}, plagiarism detection \cite{RePEc:hin:jnljam:6662984}, etc. Therefore, CRE has been attracting much interest lately.

\subsection{Contrastive Learning} 
Since the introduction of contrastive learning \cite{DBLP:conf/cvpr/ChopraHL05}, many variants have been developed. InfoNCE \cite{DBLP:journals/corr/abs-1807-03748} aims to find a positive in a group of noise samples. MoCo \cite{DBLP:conf/cvpr/He0WXG20, DBLP:journals/corr/abs-2003-04297} provides a queue-based framework for utilizing data from previous batches with the addition of a momentum encoder. Due to the use of large Transformer models \cite{DBLP:conf/naacl/DevlinCLT19, DBLP:journals/corr/abs-1907-11692, DBLP:conf/nips/YangDYCSL19} in our method, the batch size is limited. MoCo provides a solution to decouple the batch size from the number of negatives. Prototypical Contrastive Learning (PCL) \cite{DBLP:conf/iclr/0001ZXH21} treats prototypes as latent variables and brings clustering into the contrastive learning paradigm. The prototypes are found using the $k$-means clustering algorithm and its ProtoNCE loss contrasts different prototypes based on the clustering results. However, the CRE task naturally exhibits a cluster structure, and, therefore, we do not need to rely on a clustering algorithm. Our method is not to be confused with Graph Contrastive Learning (GCL) \cite{DBLP:conf/nips/YouCSCWS20,zhang2022costa} whose main inputs are graphs, whereas the inputs are text documents in this work.

%% file: sections/method.tex
\section{Our Method}
\label{sec:our_method}

\subsection{Concept Relatedness Estimation}
\label{subsec:cre}
The concept relatedness estimation (CRE) task is to predict whether two given concepts are related or unrelated. Thus, it is a binary classification task with two labels ``related'' and ``unrelated''. Formally, given a set of concepts $X = \{x_i\mid i\in \mathbb{N}_n \}$ with an index set $\mathbb{N}_n = \{1, \dots, n\}$ and a set of known pairwise binary labels $Y = \{y_{i,j} \mid  (i,j) \subset \mathbb{N}_n \times \mathbb{N}_n\}$, the CRE task is to learn a parameterized encoder $f_{\theta}$ that can estimate the true relatedness $y_{i,j}$ of two concepts $x_i$ and $x_j$. In this paper, we focus on concepts that are long documents.


\paragraph{CRE Properties.} The CRE task exhibits some unique intrinsic properties that are rarely present in typical NLP tasks. To state these properties formally, we assume three concepts $x_A$, $x_B$, and $x_C$. The similarity symbol ``$\sim$'' is used to denote that two concepts are ``related'', while the dissimilarity symbol ``$\nsim$'' is used to connect two unrelated concepts.


\begin{property}[Reflexivity]
\label{ppt:reflexivity}
A concept $x_A$ is related to itself: $x_A \sim x_A$.
\end{property}

\begin{property}[Commutativity of Relatedness]
\label{ppt:commutativity_relatedness}
If $x_A$ and $x_B$ are related, then $x_B$ and $x_A$ are related: $x_A \sim x_B \iff x_B \sim x_A$.
\end{property}

\begin{property}[Commutativity of Unrelatedness]
\label{ppt:commutativity_unrelatedness}
If $x_A$ and $x_B$ are unrelated, then $x_B$ and $x_A$ are unrelated: $x_A \nsim x_B \iff x_B \nsim x_A$.
\end{property}

\begin{property}[Transitivity of Relatedness]
\label{ppt:transitivity_relatedness}
If $x_A$ is related to $x_B$ and $x_B$ is related to $x_C$, then $x_A$ and $x_C$ are related: $(x_A \sim x_B) \land (x_B \sim x_C) \Longrightarrow x_A \sim x_C$.
\end{property}

\begin{property}[Transitivity of Unrelatedness]
\label{ppt:transitivity_unrelatedness}
If $x_A$ is related to $x_B$ but $x_B$ is unrelated to $x_C$, then $x_A$ and $x_C$ are unrelated: $(x_A \sim x_B) \land (x_B \nsim x_C) \Longrightarrow x_A \nsim x_C$.
\end{property}




\paragraph{ConcreteGraph.} To make use of the CRE properties in practice, we can build a graph to encode both the pairwise relationships from the dataset and the higher-order relationships based on those properties. We name it ConcreteGraph (\textbf{\underline{Conc}}ept \textbf{\underline{re}}la\textbf{\underline{te}}dness \textbf{\underline{Graph}}). An illustration of how to construct the ConcreteGraph is shown in Figure~\ref{fig:overview}(a). Each node in ConcreteGraph represents a concept, and edges are only added between related concept pairs. The construction of the ConcreteGraph takes $O(|Y|)$ time as we need to iterate over all labeled concept pairs. Since it is a binary classification task, we do not need to set the edge weights. By the nature of CRE, the ConcreteGraph usually has multiple connected components, where each component contains a set of related concepts. To identify the components of the ConcreteGraph, we can use breadth-first search (BFS), which takes $O(V + E)$ time where $V$ is the number of nodes and $E$ is the number of edges.

\subsection{Data Augmentation} 
\label{subsec:data_aug}
The ConcreteGraph enables a straightforward data augmentation method for addressing the data scarcity problem. The CRE properties automatically come into play when we sample new concept pairs from the ConcreteGraph. \textit{Commutativity} is used if we sample two nodes that are already provided by the dataset but in a different order. For example, if $(x_A, x_B)$ is provided and we sample $(x_B, x_A)$, then this new pair is validated by the \textit{commutativity of relatedness} property (Property~\ref{ppt:commutativity_relatedness}) or the \textit{commutativity of unrelatedness} property (Property~\ref{ppt:commutativity_unrelatedness}). \textit{Transitivity} is utilized when the path between the two sampled concepts has at least two edges or when there is no path between them. For example, if there is a path $x_A$\textemdash$x_B$\textemdash$x_C$, then the new concept pair $(x_A,x_C)$ is justified by the \textit{transitivity of relatedness} property (Property~\ref{ppt:transitivity_relatedness}). When two concepts are sampled from two different components, the \textit{transitivity of unrelatedness} property (Property~\ref{ppt:transitivity_unrelatedness}) is employed to prove that they are unrelated.

In theory, any two concepts within the same graph component are related; any two concepts from two different components are unrelated. But it is intractable to use all possible concept pairs. For example, the largest component of the ConcreteGraph of the WORD dataset \cite{DBLP:conf/lrec/Ein-DorHKLMRSS18} has 4,301 nodes. We could sample up to 9,247,150 related concept pairs from it, which is impractical for a Transformer model to learn. Therefore, we must only use a subset of those concept pairs. An intuitive heuristic is to pick a concept pair with a short path length. If the path length is too long, the connection between the concept pair becomes ``risky''. Namely, when there are more edges on the path, the probability of the existence of a noisy edge becomes higher, which could degrade the quality of the path. We later prove this theory with an empirical ablation study in Sub-section~\ref{subsec:aug_ablation}. In practice, we obtain the $k$-hop neighborhood for every node, where $k$ is a small integer, such as 2 or 3. We use a subset of the nodes in this local neighborhood as related concepts by setting a target augmentation ratio. Unrelated concept pairs can be produced by simply sampling concepts from other components.

\subsection{Graph Component Contrastive Learning}
\label{subsec:gccl}
We introduce Graph Component Contrastive Learning (GCCL) to take full advantage of local and global structural information of the ConcreteGraph, which is neglected by previous CRE methods. Our ConcreteGraph-based data augmentation only provides concept pairs from a local $k$-hop neighborhood, and it is intractable to include all concept pairs that satisfy the CRE properties. In fact, it is harmful to explicitly extract all possible concept pairs as augmented data due to the quality problem caused by long paths. GCCL captures global structural information by learning the representations of the components in the ConcreteGraph and contrasting a concept's own component against other components. In this way, all of the related and unrelated concept pairs are preserved implicitly because each component embedding aggregates a group of related concept embeddings. Our method also has a connection to clustering methods: if we add edges for all possible related pairs, the ConcreteGraph itself would become a cluster graph because all the components become complete subgraphs. For that reason, a ConcreteGraph component in GCCL is equivalent to a cluster (prototype) in Prototypical Contrastive Learning (PCL) \cite{DBLP:conf/iclr/0001ZXH21}. The difference is that GCCL does not rely on clustering algorithms to find clusters.

The overview of our GCCL framework is illustrated in Figure~\ref{fig:overview}(b). The main objective of GCCL is to treat the embedding of a concept's own component embedding as the positive and the embeddings of all other components as negatives. This allows the model to learn representations for both individual concepts and their components, which implicitly encode the global graph structure. Formally, we assume that the ConcreteGraph has a set of $r$ components $C = \{c_1, c_2, \dots, c_r\}$, where each component itself is a set of concepts $c_s = \{x_{s_1}, x_{s_2}, \dots, x_{s_p}\}$ and $p$ is the number of concepts in $c_s$. Same as PCL, we assume that $f_{\theta}(x_i)$ follows an isotropic Gaussian distribution around the component embedding:
\begin{equation} \label{eq:gaussian}
\resizebox{.9 \columnwidth}{!}{
$p(x_i; c_s, \theta) = \frac{\exp \big( -\big(f_{\theta}(x_i) - f_{\theta}(c_s)\big)^2 /2\sigma_s^2 \big)}{\sum_{j=1}^{r}\exp \big( -\big(f_{\theta}(x_i) - f_{\theta}(c_j) \big)^2/2\sigma_j^2 \big)},$
}
\end{equation}
where $x_i \in c_s$ and $\sigma^2$ is the variance. We also overload $f_{\theta}$ with the input of a component, such that we get the representation of a component using a shared backbone encoder. In this paper, the component embedding is simply the element-wise average of the embeddings of its concepts: $f_{\theta}(c_s) = \frac{1}{p}\sum_{k=1}^pf_{\theta}(x_{s_k})$. Theoretically, we could add an extra Graph Neural Network (GNN) layer \cite{DBLP:journals/tnn/WuPCLZY21} on top of the concept embeddings to get the component embedding. However, because the components are complete subgraphs, GNN layers would quickly over-smooth the concept embeddings, which is close to their average. Thus, we do not use GNN layers to obtain the component embeddings.


After applying $\ell_2$-normalization to both the concept embeddings and the component embeddings, we have $\big(f_{\theta}(x_i) - f_{\theta}(c_s) \big)^2 = 2 - 2f_{\theta}(x_i) \cdot f_{\theta}(c_s)$. Then the maximum log-likelihood estimation for $\theta$ optimizes a novel Graph-Component-NCE (GC-NCE) loss $\mathcal{L}_{\text{GC-NCE}}$:
\begin{align}
\nonumber
\theta^* ={}& \argmax_{\theta}\sum_{i=1}^{n} \log p(x_i;C, \theta) \notag\\
={}&\argmin_{\theta} \sum_{i=1}^{n} -\log \frac{\exp \big( f_{\theta}(x_i) \cdot f_{\theta}(c_s)/\phi_s \big)}{\sum_{j=1}^{r}\exp \big( f_{\theta}(x_i) \cdot f_{\theta}(c_j)/\phi_j \big)}  \notag\\
\triangleq{}& \argmin_{\theta} \mathcal{L}_{\text{GC-NCE}},
\label{eq:gc-nce}
\end{align}
where 
\begin{equation} \label{eq:mll}
\begin{split}
\phi_s = \frac{\sum_{k=1}^p \norm{ f_{\theta}(x_{s_k}) - f_{\theta}(c_s) }_{2}}{p\log(p+\alpha)}
\end{split}
\end{equation}
is a concentration estimation that replaces the variance in Eq.~\ref{eq:gaussian}, which is also introduced in PCL; $\alpha$ is a smoothing hyper-parameter for preventing $\phi$ from becoming excessively large when the component size $p$ is small. When $\phi$ is small, the concept embeddings within the same component are pulled closer together and, thus, have a high concentration level.

In addition to the main GC-NCE objective of GCCL, we also use the InfoNCE \cite{DBLP:journals/corr/abs-1807-03748} objective under the MoCo framework \cite{DBLP:conf/cvpr/He0WXG20, DBLP:journals/corr/abs-2003-04297}. The idea of InfoNCE is to distinguish the embedding of a concept from the embeddings of a set of noise concepts. MoCo introduces a dynamic dictionary based on a queue data structure for storing the embeddings from a momentum encoder $f_{\theta_k}$ in a first-in-first-out (FIFO) fashion, which enables the decoupling between the batch size and the amount of negatives. This is especially useful since we use large Transformer models as encoders. The MoCo objective is constructed by contrasting the query embedding $f_{\theta}(x)$ from the main encoder and the key embeddings $f_{\theta_k}(x)$ from the momentum encoder. The momentum encoder $f_{\theta_k}$ is the moving average of the main encoder $f_{\theta}$ with a momentum coefficient $m\in [0, 1)$:
\begin{equation} \label{eq:momentum}
\begin{split}
\theta_k \leftarrow m\theta_k + (1-m)\theta_q.
\end{split}
\end{equation}

Similar to InfoNCE, the MoCo loss treats the query and key embeddings of a concept as the positive pair, and the key embeddings of other concepts in the MoCo queue are negatives to the query embedding of the current concept. That leads to the MoCo objective:
\begin{equation} \label{eq:moco}
\begin{split}
\mathcal{L}_{\text{MoCo}} = \sum_{i=1}^{n}-\log \frac{ \exp\big( f_{\theta}(x_i) \cdot f_{\theta_k}(x_i) / \tau \big) }{ \sum_{j=0}^{Q}\exp \big( f_{\theta}(x_i) \cdot f_{\theta_k}(x_j) / \tau \big) },
\end{split}
\end{equation}
where the denominator includes one positive and $Q$ negatives in the MoCo queue; $\tau$ is a temperature hyper-parameter. The purpose of adding this additional MoCo objective is to preserve local smoothness of the loss function \cite{DBLP:conf/iclr/0001ZXH21}: the GC-NCE objective (Eq.~\ref{eq:gc-nce}) and the CRE objective (Eq.~\ref{eq:cre}) both pull the embeddings of related concepts closer, those concept embeddings might become very similar after finetuning, which can cause poor generalization; the MoCo objective preserves the distinction of individual concept embeddings by contrasting a concept against all other concepts.


In this paper, we use Transformer \cite{DBLP:conf/nips/VaswaniSPUJGKP17} as the encoder $f_{\theta}$. After passing the concepts into $f_{\theta}$ and obtaining their concept embeddings, the concept relatedness is estimated as follows:
\begin{equation} \label{eq:cre}
\begin{split}
h(x_i, x_j; \theta) & = f_{\theta}(x_i) \oplus f_{\theta}(x_j) \oplus \big|f_{\theta}(x_i) - f_{\theta}(x_j) \big| \\
p(x_i, x_j) & = \sigma \big(\mathbf{W}h(x_i, x_j; \theta)  \big)\\
\mathcal{L}_{\text{CRE}} & = L_{\text{CE}}\big( p(x_i, x_j), y_{i,j} \big),
\end{split}
\end{equation}
where $\oplus$ is the concatenation operation of vectors; $\big| f_{\theta}(x_i) - f_{\theta}(x_j) \big|$ is the element-wise absolute value of the difference of the two vectors; $\sigma$ is a sigmoid activation function; $\mathbf{W}$ is a trainable parameter matrix. We use binary cross-entropy loss $L_{\text{CE}}$ as the loss function.

The overall objective $\mathcal{L}$ combines the CRE loss (Eq.~\ref{eq:cre}), the GC-NCE loss (Eq.~\ref{eq:gc-nce}) and the MoCo loss (Eq.~\ref{eq:moco}):
\begin{equation} \label{eq:method-objective}
\begin{split}
\mathcal{L} &= \mathcal{L}_{\text{CRE}} + \beta \mathcal{L}_{\text{GCCL}} \\
&= \mathcal{L}_{\text{CRE}} + \beta (\mathcal{L}_{\text{GC-NCE}} + \mathcal{L}_{\text{MoCo}}),
\end{split}
\end{equation}
where $\beta$ is the hyper-parameter that adjusts the effect of the GCCL loss $\mathcal{L}_{\text{GCCL}}$.


%% file: sections/experiments.tex
\section{Experiments}
\label{sec:experiments}
We conduct experiments on three CRE datasets: WORD \cite{DBLP:conf/lrec/Ein-DorHKLMRSS18}, CNSE and CNSS \cite{DBLP:conf/acl/LiuNWGHLX19}. We compare the performance of three Transformer models, including BERT \cite{DBLP:conf/naacl/DevlinCLT19}, RoBERTa \cite{DBLP:journals/corr/abs-1907-11692} and XLNet \cite{DBLP:conf/nips/YangDYCSL19}. We also include four latest Graph Neural Network (GNN) models \cite{DBLP:conf/icml/ChenWHDL20, DBLP:conf/nips/ZengZXSMKPJC21, DBLP:conf/icml/WangZ22, DBLP:conf/acl/LiuNWGHLX19} as baselines. Detailed ablation studies show the effect of different modules of our method.

\begin{table*}[tb]
  \centering
  \resizebox{2\columnwidth}{!}{\begin{tabular}{l|c c c|c c c|c c c}
    \toprule
        & \multicolumn{3}{c|}{\textbf{WORD}}  & \multicolumn{3}{c|}{\textbf{CNSE}}  & \multicolumn{3}{c}{\textbf{CNSS}}  \\ 
    Model & Acc & F1 & AUC & Acc & F1 & AUC & Acc & F1 & AUC \\ 

    \Xhline{2\arrayrulewidth}

    GCNII \cite{DBLP:conf/icml/ChenWHDL20} & 54.62 & 44.55 & 57.03 &  62.81 & 56.22 & 66.40 & 58.40 & 61.06 & 63.00 \\
    ShaDow-GNN \cite{DBLP:conf/nips/ZengZXSMKPJC21} & 62.27 &  65.96 & 67.39 & 60.78 & 57.34 & 64.75 & 60.40 & 67.63 & 64.85\\
    JacobiConv \cite{DBLP:conf/icml/WangZ22} & 62.13 & 60.30 & 66.89 & 68.32 & 59.69 & 76.42 & 62.94 & 66.43 & 67.87 \\
    GCN / CIG \cite{DBLP:conf/acl/LiuNWGHLX19} & 62.81 & 67.75 & 67.00 & 83.95   & 82.05  & 91.92 & 89.69 & 90.05 & 95.97  \\
    \hspace{2em} (Corrected CIG) & - & - & -    & (77.57) & (75.13) & -     & (82.87) & (83.43) & -  \\

    \Xhline{2\arrayrulewidth}

    BERT \cite{DBLP:conf/naacl/DevlinCLT19} & 76.85 & 76.12 & 84.86 & 83.97 & 81.67 & 92.10 & 90.63 & 91.09 & 97.07 \\
    \hspace{2em} w. Aug \hspace{3.5em} (\textbf{Ours}) & 77.42 & 76.81 & 85.75 & 85.17 & 83.39 & 93.08 & 91.32 & 91.71 & 97.50 \\
    \hspace{2em} w. GCCL \hspace{2.5em} (\textbf{Ours}) & 77.83 & 77.59 & 85.86 & \textbf{85.79} & 84.18 & 92.96 & \textbf{93.67} & \textbf{93.78} & \textbf{98.37} \\
    \hspace{2em} w. Aug + GCCL (\textbf{Ours}) & \textbf{78.18} & \textbf{78.25} & * \textbf{86.60} & 85.76 & \textbf{84.84} & \textbf{93.46} & 93.32 & 93.54 & 98.16 \\

    \hline

    RoBERTa \cite{DBLP:journals/corr/abs-1907-11692} & 77.25 & \textbf{78.11} & 85.55 & 84.52 & 82.79 & 92.88 & 92.84 & 92.74 & 98.20 \\
    \hspace{2em} w. Aug \hspace{3.5em} (\textbf{Ours}) & 77.88 & 76.34 & 86.30 & 85.35 & 84.20 & 93.97 & 93.08 & 93.19 & 98.11 \\
    \hspace{2em} w. GCCL \hspace{2.5em} (\textbf{Ours}) & \textbf{78.61} & 77.93 & 86.26 & 86.21 & 84.40 & 93.94 & \textbf{93.82} & \textbf{93.93} & \textbf{98.50} \\
    \hspace{2em} w. Aug + GCCL (\textbf{Ours}) & 78.14 & 76.69 & \textbf{86.45} & \textbf{87.00} & \textbf{86.05} & \textbf{94.39} & 93.46 & 93.62 & 98.05 \\

    \hline

    XLNet \cite{DBLP:conf/nips/YangDYCSL19} & 77.18 & * \textbf{78.42} & 86.10 & 87.00 & 84.65 & 95.21 & 93.70 & 93.84 & 98.53 \\
    \hspace{2em} w. Aug \hspace{3.5em} (\textbf{Ours}) & 77.75 & 76.51 & 86.00 & 88.41 & 86.87 & 95.29 & 93.79 & 93.84 & 98.40 \\
    \hspace{2em} w. GCCL \hspace{2.5em} (\textbf{Ours}) & 77.98 & 77.56 & 85.72 & 88.06 & 86.47 & 95.21 & * \textbf{94.60} & * \textbf{94.74} & * \textbf{98.80} \\
    \hspace{2em} w. Aug + GCCL (\textbf{Ours}) & * \textbf{79.15} & 78.32 & \textbf{86.25} & * \textbf{88.82} & * \textbf{87.70} & * \textbf{95.53} & 94.15 & 94.34 & 98.68 \\
    \bottomrule
  \end{tabular}}
  \caption{Performance comparison with various baselines for our ConcreteGraph-based data augmentation method and Graph Component Contrastive Learning (GCCL) framework. The best results are highlighted with bold text for each model-dataset combination; the best results for each of the three datasets are marked with *.}
  \label{tb:performance_comp}
\end{table*}

\subsection{Datasets}

The Wikipedia Oriented Relatedness Dataset (WORD) dataset \cite{DBLP:conf/lrec/Ein-DorHKLMRSS18} collects English concepts from Wikipedia, including 19,176 concept pairs. The Chinese News Same Event (CNSE) dataset and the Chinese News Same Story (CNSS) dataset were both created by \citet{DBLP:conf/acl/LiuNWGHLX19}, containing 29,063 and 33,503 news article pairs from the major internet news providers in China, respectively. We use the official dataset split for WORD whose train-test ratio is approximately 2:1. Since CNSE and CNSS do not provide an official dataset split, they are split randomly with a train-dev-test ratio of 7:2:1. These dataset splits are fixed throughout different experiments across all models.


\subsection{Experimental Setup}
We use the AdamW optimizer \cite{DBLP:conf/iclr/LoshchilovH19} with learning rate $= 1 \times 10^{-5}$ and $\epsilon = 1 \times 10^{-8}$, following a linear schedule. The Transformer models are trained for 5 epochs. For GC-NCE, we use $\alpha = 10$. For MoCo \cite{DBLP:conf/cvpr/He0WXG20, DBLP:journals/corr/abs-2003-04297}, we use queue size $Q=32$, momentum coefficient $m=1-1 \times 10^{-4}$, and temperature $\tau = 0.1$. We use $\beta=0.1$ for the overall loss. CIG \cite{DBLP:conf/acl/LiuNWGHLX19} is trained with its default hyper-parameters. Experiments are conducted on four Nvidia TITAN V GPUs.

\subsection{Performance Comparison}
\label{subsec:comparison}
The performance comparison of different models is summarized in Table~\ref{tb:performance_comp}. Three performance metrics are used: accuracy, F1, and AUC. We compare our method with different GNNs and Transformers.


We use four recent GNNs as baselines. For GCNII \cite{DBLP:conf/icml/ChenWHDL20}, ShaDow-GNN \cite{DBLP:conf/nips/ZengZXSMKPJC21} and JacobiConv \cite{DBLP:conf/icml/WangZ22}, we build a complete graph for each concept, where each node is a token and the node features are BERT token embeddings. For CNSE and CNSS, Concept Interaction Graph (CIG) \cite{DBLP:conf/acl/LiuNWGHLX19} is the current state-of-the-art model. CIG extracts a concept graph for each article, where its concepts are different from CRE concepts in that they are more similar to named entities. One major problem with this model is that the size of the concept graphs is limited. If a concept graph exceeds the limit, the model simply discards the article pair. Therefore, CIG practically uses an easier subset of the original dataset, causing inaccurate measurement of its performance. Thus, we add a ``Corrected CIG'' row in the table to include the corrected accuracy and F1. Because CIG is based on a graph convolutional network (GCN) \cite{Fey/Lenssen/2019} but it only works on CNSE and CNSS, we developed a similar GCN based on ELMo \cite{DBLP:conf/naacl/PetersNIGCLZ18} for the WORD dataset. Because most GNNs are designed to process sparse graph inputs and they can suffer from the over-smoothing issue, they usually do not excel at processing text inputs.



We finetune three Transformer models, BERT \cite{DBLP:conf/naacl/DevlinCLT19}, RoBERTa \cite{DBLP:journals/corr/abs-1907-11692}, and XLNet \cite{DBLP:conf/nips/YangDYCSL19} on the three datasets. GCCL is independent of the ConcreteGraph-based data augmentation method. Thus, they can be used either separately or simultaneously. To illustrate their benefits, we report the performance of the base Transformer model, Transformer with data augmentation (``w. Aug''), Transformer with GCCL (``w. GCCL''), and Transformer with both modules (``w. Aug + GCCL''). 


When compared with GNNs, Transformer models exhibit a large advantage because the self-attention mechanism automatically optimizes the attention weights between tokens, whereas the adjacency matrices in GNNs are fixed. In addition, the base Transformer models can even outperform the uncorrected CIG and, with the help of our GCCL framework, the margin is enlarged. When comparing our methods with the base Transformer models, we can see that the GCCL framework is able to bring more improvement than our data augmentation method in most cases. This shows that the global structural information from GCCL is indeed more effective for the CRE task. 

When ``Aug'' and ``GCCL'' are combined, the performance is further enhanced in most metrics on WORD and CNSE; however, applying only GCCL produces better results on the CNSS dataset. ConcreteGraph-based data augmentation is confined to a local $k$-hop neighborhood, while GCCL captures the information of the entire component. They can complement each other in most cases but their difference may also cause disagreement as to whether two concepts are related if they are connected by a long path. That is, data augmentation does not sample any concept pairs with path lengths longer than $k$, but GCCL retains those connections. Such contradiction sometimes worsens the performance.


\subsection{Ablation Study of GCCL}
\label{subsec:gccl_ablation}
The GCCL loss has two parts: the main GC-NCE loss and the auxiliary MoCo loss. We conduct an ablation study on their individual effect on BERT \cite{DBLP:conf/naacl/DevlinCLT19}. As shown in Table~\ref{table:gccl_ablation}, both the MoCo loss and the GC-NCE loss can improve the model performance. In addition, the main GC-NCE objective produces more performance improvement than MoCo does. When they are combined, the performance improvement from the two objectives stacks. 

\begin{table}[t]
 \centering
 \resizebox{1\columnwidth}{!}{%
    \begin{tabular}{ l | c   c   c | c   c   c  | c   c   c }
    \toprule
        & \multicolumn{3}{c|}{\textbf{WORD}}  & \multicolumn{3}{c|}{\textbf{CNSE}}  & \multicolumn{3}{c}{\textbf{CNSS}}  \\ 
                & Acc & F1 & AUC & Acc & F1 & AUC & Acc & F1 & AUC \\ 
        \Xhline{2\arrayrulewidth}
        BERT    & 76.85 & 76.12 & 84.86 & 83.97 & 81.67 & 92.10 & 90.63 & 91.09 & 97.07\\
        w. MoCo & 77.10 & 76.79 & 84.82 & 84.07 & 81.78 & 92.25 & 92.57 & 92.88 & 97.96\\
        w. GC-NCE & 77.58 & \textbf{78.06} & 85.81 & 84.93 & 83.17 & 92.68 & 92.93 & 93.06 & 98.15 \\
        w. GCCL & \textbf{77.83} & 77.59 & \textbf{85.86} & \textbf{85.79} & \textbf{84.18} & \textbf{92.96} & \textbf{93.67} & \textbf{93.78} & \textbf{98.37}\\
     \bottomrule
    \end{tabular}
 }
 \caption{The ablation study of the GCCL objective.}
 \label{table:gccl_ablation}
\end{table}

\begin{table}[t]
 \centering
 \small
 \resizebox{1.0\columnwidth}{!}{%
    \begin{tabular}{ c | c | c | c | c | c | c | c }
    \toprule
        $k$-hop & \thead{Realized\\Aug\\Ratio} & \thead{Target\\Aug\\Ratio} & \thead{Sampled\\Related\\Pairs} & \thead{Possible\\Related\\Pairs} & Accuracy & F1 & AUC\\
        \Xhline{2\arrayrulewidth}
        1 & - & - & 6,842 & - & 76.85 & 76.12 & 84.86\\\hline
        2 & $\times$ 2 & $\times$ 2 & 12,875 & 106,769 & \textbf{77.42} & \textbf{76.81} & \textbf{85.75} \\
        
        \Xhline{2\arrayrulewidth}
        
          & $\times$ 3 & $\times$ 3 & 19,312 & 106,769 & 75.66 & 74.51 & 83.96 \\
        2 & $\times$ 4 & $\times$ 4 & 25,750 & 106,769 & 72.23 & 72.07 & 79.99 \\
          & $\times$ 5 & $\times$ 5 & 32,187 & 106,769 & 70.99 & 70.78 & 78.95 \\
        
        \Xhline{2\arrayrulewidth}
        
        2  & $\times$ 16.59 & $\infty$ & 106,769 & 106,769 & 63.51 & 58.65 & 70.97\\
        3 &  $\times$ 62.28 & $\infty$ & 400,923 & 400,923 & 56.76 & 51.38 & 63.69\\

        \Xhline{2\arrayrulewidth}
        
        3  & $\times$ 2 & $\times$ 2 & 12,875 & 400,923 & 75.91 & 75.81 & 83.78\\
        4  & $\times$ 2 & $\times$ 2 & 12,875 & 1,428,758 & 74.90 & 75.44 & 83.08\\
        5  & $\times$ 2 & $\times$ 2 & 12,875 & 3,936,722 & 74.06 & 75.17 & 81.56\\
    \bottomrule
    \end{tabular}
 }
 \caption{The effect of the number of hops $k$ and augmentation ratio in the ConcreteGraph-based data augmentation method. The number of sampled unrelated pairs is equal to the number of sampled related pairs and it is omitted.}
 \label{table:data_aug_ablation}
\end{table}

\begin{figure}[t]
\centering
\begin{subfigure}{0.7\columnwidth}
  \centering
  \includegraphics[width=1\columnwidth,trim={5cm 2cm 4cm 1cm},clip]{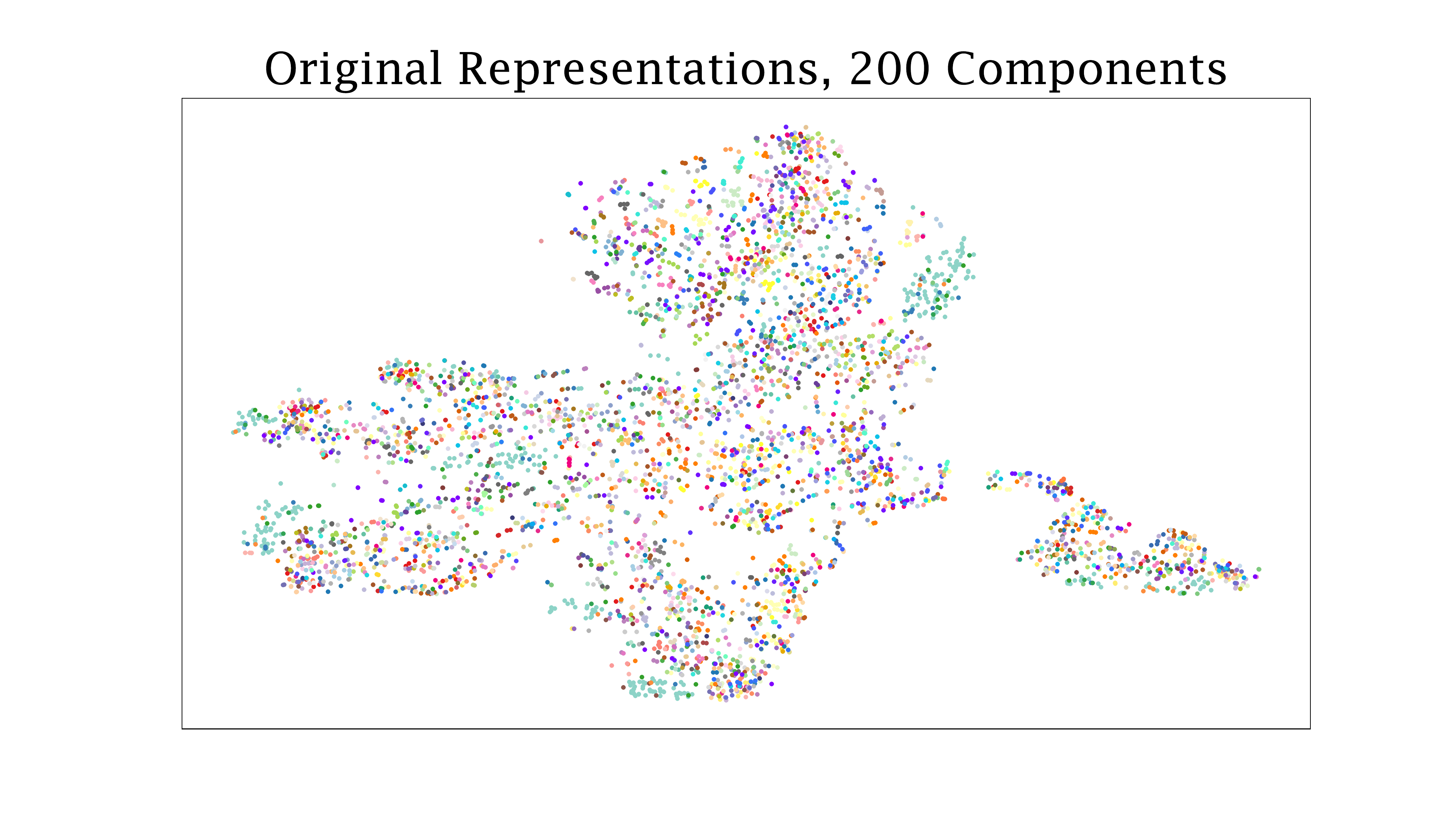}
  \label{fig:sub-first}
\end{subfigure}
\begin{subfigure}{0.7\columnwidth}
  \centering
  \includegraphics[width=1\columnwidth,trim={5cm 2cm 4cm 1cm},clip]{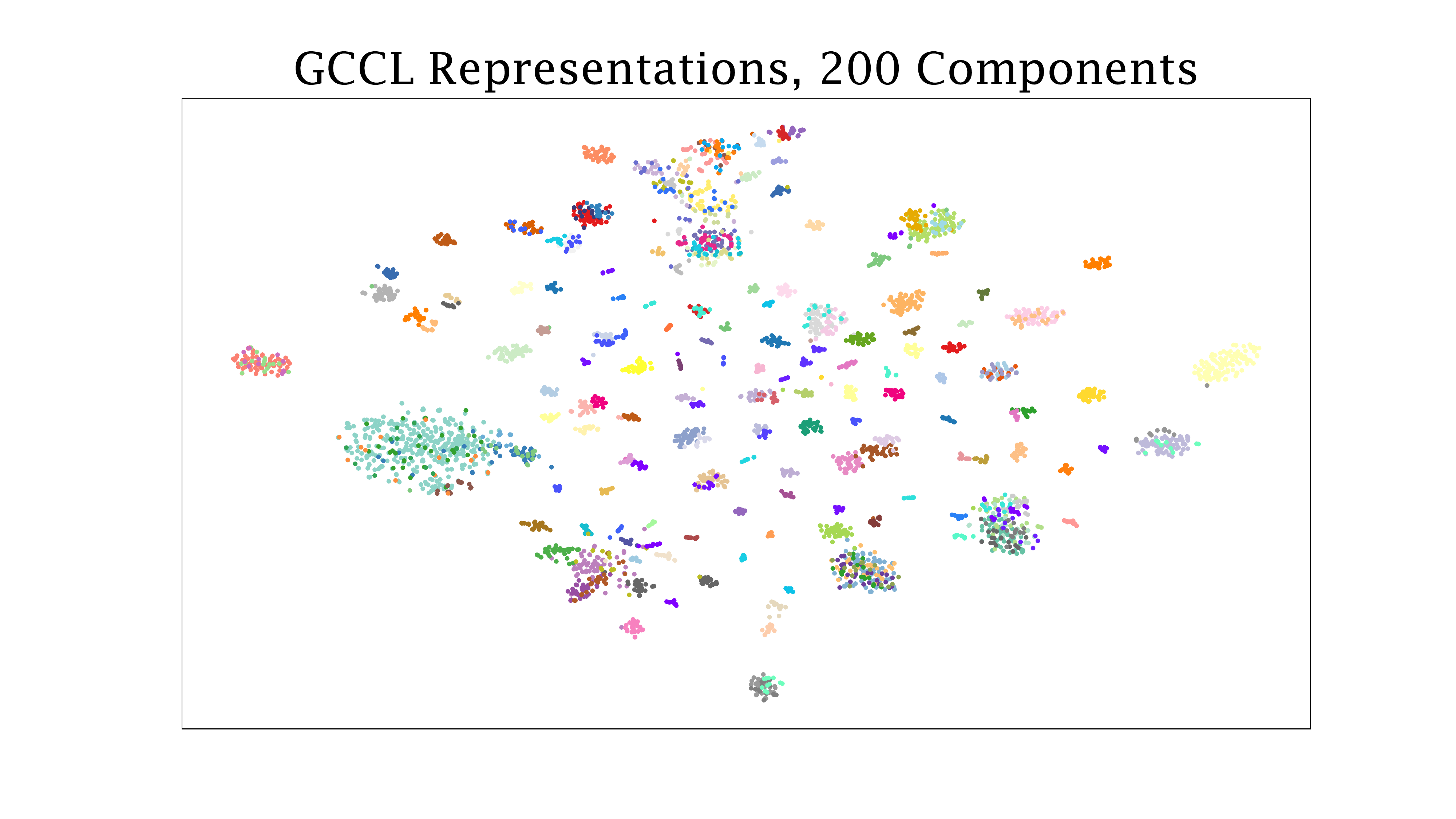}
  \label{fig:sub-third}
\end{subfigure}

\caption{The t-SNE visualization of the original and GCCL concept embeddings. Each point corresponds to a concept embedding, and the concepts in the same component share the same color.}
\label{fig:tsne-visual}
\end{figure}

In Figure~\ref{fig:tsne-visual}, we also visualize the concept embeddings with t-SNE \cite{van2008visualizing} to qualitatively illustrate the effect of GCCL. GCCL better represents the concepts by taking the global structure of the ConcreteGraph into account: the embeddings of the concepts in the same component are closer to each other using GCCL and the embedding space of GCCL is better organized than that of the original model. This gives us an insight into how GCCL helps the main CRE objective: GCCL pulls related concepts together and separates unrelated components. 

\subsection{Ablation Study of Data Augmentation}
\label{subsec:aug_ablation}
To understand whether the quality or the quantity of augmented data better helps the model performance, we experiment with different $k$ values and target augmentation ratios, as summarized in Table~\ref{table:data_aug_ablation}. When the target augmentation ratio is specified, the sampling process of new related concept pairs stops when the target augmentation ratio is reached, regardless of whether all possible related concept pairs are sampled. If there is no target augmentation ratio ($\infty$), we find all possible related concept pairs for the given $k$. In both cases, we keep sampling unrelated concept pairs until the number of unrelated pairs is equal to that of related pairs. This ensures the balance of the augmented dataset.

When $k$ is fixed at 2 and the augmentation ratio increases, the performance worsens. If we take it to the extreme by not setting a target ratio, the performance drops significantly. These results show that a large augmentation ratio does not always help because it shrinks the proportion of annotated concept pairs and the quality of sampled concept pairs never matches that of annotated concept pairs. When the augmentation ratio is fixed at 2 and $k$ is incremented, the performance also declines. This confirms our theory about how path length can affect the quality of sampled concept pairs. Namely, when the path length between two concepts is long, the existence of a noisy edge becomes more likely, which degrades the overall quality of the connection.